\documentclass{article}
\PassOptionsToPackage{table,xcdraw,dvipsnames}{xcolor}
\usepackage{float}

\usepackage{microtype}
\usepackage{graphicx}
\usepackage{subcaption}
\usepackage{booktabs} 
\usepackage{tcolorbox}
\usepackage{hyperref}
\usepackage{makecell}

\usepackage[accepted]{icml2025}

\usepackage{amsmath}
\usepackage{amssymb}
\usepackage{mathtools}
\usepackage{amsthm}
\usepackage{makecell}
\usepackage{multirow}
\usepackage[capitalize,noabbrev]{cleveref}

\theoremstyle{plain}

\theoremstyle{definition}

\theoremstyle{remark}

\usepackage[textsize=tiny]{todonotes}

\usepackage{url}

\icmltitlerunning{MASSV: Multimodal Adaptation and Self-Data Distillation for Speculative Decoding of Vision-Language Models}

\begin{document}

\twocolumn[
\icmltitle{MASSV: Multimodal Adaptation and Self-Data Distillation for \\ Speculative Decoding of Vision-Language Models}



\icmlsetsymbol{equal}{*}

\begin{icmlauthorlist}
\icmlauthor{Mugilan Ganesan$^{\text{*}}$}{yyy}
\icmlauthor{Shane Segal}{yyy}
\icmlauthor{Ankur Aggarwal}{yyy}
\icmlauthor{Nish Sinnadurai}{yyy}
\icmlauthor{Sean Lie}{yyy}
\icmlauthor{Vithursan Thangarasa}{yyy}
\end{icmlauthorlist}

\icmlaffiliation{yyy}{Cerebras Systems, Sunnyvale, California}

\icmlcorrespondingauthor{Mugilan Ganesan}{mugilan.ganesan@cerebras.net}
\icmlcorrespondingauthor{Vithursan Thangarasa}{vithu@cerebras.net}

\icmlkeywords{Machine Learning, ICML}

\vskip 0.3in
]



\printAffiliationsAndNotice{\icmlEqualContribution} 

\begin{figure*}[h!]
    \centering
    \vskip -0.1in
    \includegraphics[width=0.9\linewidth]{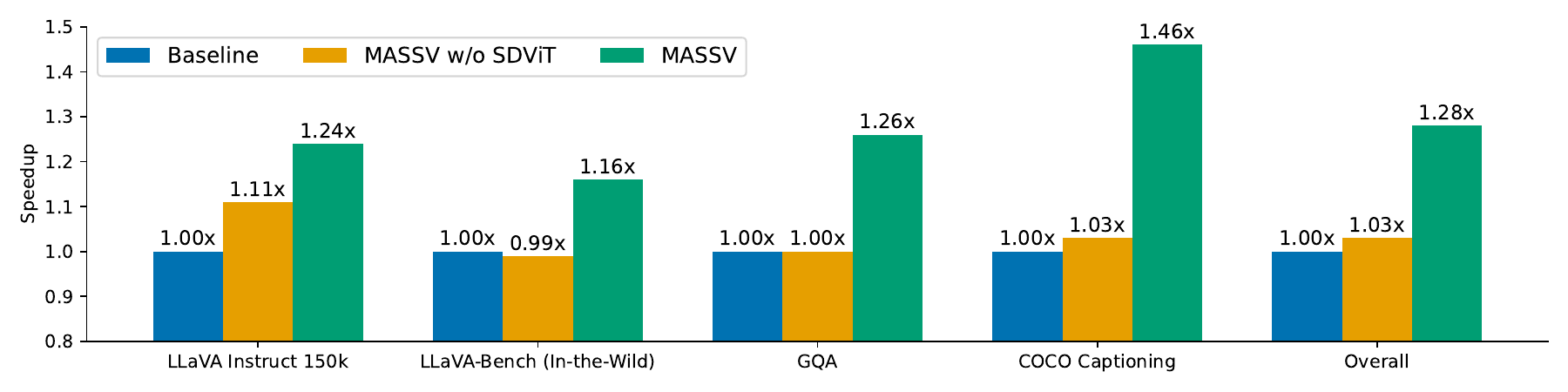}
    \vskip -0.1in
\caption{End-to-end wallclock speedups when drafting for Qwen2.5-VL 7B Instruct at temperature $T = 0$ with speculation length $\gamma = 5$. The baseline uses Qwen2.5-1.5B as a text-only drafter (image tokens removed). MASSV consistently yields the highest speedups across all categories, achieving up to 1.46$\times$ on COCO captioning and 1.28$\times$ overall. The gains are most pronounced for visually grounded tasks, demonstrating the importance of multimodal adaptation and self-distilled visual instruction for accelerating VLM inference.}

    \label{fig:speedup_qwen}
\end{figure*}

\begin{abstract}
Speculative decoding significantly accelerates language model inference by
enabling a lightweight draft model to propose multiple tokens that a larger
target model verifies simultaneously. However, applying this technique to
vision-language models (VLMs) presents two fundamental challenges: small
language models that could serve as efficient drafters lack the architectural
components to process visual inputs, and their token predictions fail to match
those of VLM target models that consider visual context. We introduce
\textbf{M}ultimodal \textbf{A}daptation and \textbf{S}elf-Data Distillation for
\textbf{S}peculative Decoding of \textbf{V}ision-Language Models (MASSV), which
transforms existing small language models into effective multimodal drafters
through a two-phase approach. MASSV first connects the target VLM's vision
encoder to the draft model via a lightweight trainable projector, then applies
self-distilled visual instruction tuning using responses generated by the target
VLM to align token predictions. Comprehensive experiments across the Qwen2.5-VL
and Gemma3 model families demonstrate that MASSV increases accepted length by up
to 30\% and delivers end-to-end inference speedups of \textbf{up to 1.46x}
compared to conventional text-only drafting baselines on visually-grounded
tasks. MASSV provides a scalable, architecture-compatible method for
accelerating both current and future VLMs.
\end{abstract}
\section{Introduction}

Large language models (LLMs) have transformed artificial intelligence by
delivering breakthrough capabilities in reasoning~\citep{jaech2024openai,
deepseekai2025deepseekr1incentivizingreasoningcapability}, code
generation~\citep{hui2024qwen25codertechnicalreport, li2023starcodersourceyou},
and natural language understanding~\citep{openai2023gpt4, geminiteam2023gemini,
claude3, grattafiori2024llama3herdmodels}. However, these achievements come with
substantial computational costs, particularly during inference. The fundamental
constraint arises from autoregressive generation, where each token must be
predicted sequentially based on all previous tokens, creating an inherent
bottleneck that limits parallelization. Speculative decoding (SD) addresses this
bottleneck by leveraging smaller draft models to generate multiple candidate
tokens autoregressively, which are then verified in parallel by the larger
target model~\citep{chen2023acceleratinglargelanguagemodel, leviathan2023}. This
technique reduces sequential operations while preserving the original output
distribution, effectively amortizing the computational cost and enabling
substantial inference speedups without quality degradation.

While SD has been well-studied for text-only models, extending it to
vision-language models (VLMs) introduces unique challenges. VLMs process
multimodal inputs by mapping image features and text tokens into a joint
embedding space, enabling sophisticated visual reasoning
capabilities~\citep{pmlr-v139-radford21a, liu2023visual}. This multimodal
conditioning presents two fundamental challenges for SD: (1) architectural
incompatibility, as small language models lack the components to process visual
inputs, and (2) distribution mismatch, as unimodal draft models cannot
effectively capture the visually-grounded nature of the target VLM's outputs.
Previous approaches have addressed these challenges either by excluding image
tokens entirely or by training small multimodal models from
scratch~\citep{gagrani2024speculativedecodingmultimodallarge}. The former
approach fails to leverage visual information, while the latter requires
substantial computational resources and may still suffer from distribution
misalignment. \citet{lee2024inbatch} explored ensemble-based methods that
combine multiple drafting strategies through batch inference, achieving
robustness across diverse input scenarios without requiring additional model
parameters. However, these ensemble approaches do not fundamentally address the
distribution mismatch between draft and target models, instead relying on
averaging predictions from multiple unaligned drafters. Neither of these
approaches fully exploit the potential of existing model families or directly
optimize for the distribution alignment needed for effective SD.

We introduce \textbf{M}ultimodal \textbf{A}daptation and \textbf{S}elf-Data
Distillation for \textbf{S}peculative Decoding of \textbf{V}ision- Language
Models (MASSV), a principled method for adapting smaller language models from
the same family as the target VLM to serve as efficient multimodal draft models.
Our approach consists of two key components. First, we formulate the multimodal
drafting problem as mapping from a target VLM's vision-language embedding space
to a draft LM's embedding space, constructing a drafter by connecting the target
VLM's vision encoder and multimodal projector to a smaller language model from
the same family. Second, we propose a training methodology centered on self-data
distillation~\citep{thangarasa2024selfdatadistillationrecoveringquality,
Yang2024SelfDistillationBD} to align the draft model's distribution with the
target model's, specifically optimizing for higher token acceptance rates during
SD. As shown in Figure~\ref{fig:speedup_qwen}, MASSV achieves significant
end-to-end speedups, particularly on visually grounded tasks (e.g., up to 1.46x
on COCO captioning), demonstrating the importance of multimodal adaptation and
self-data distillation for improving acceptance rate of draft tokens. Our
contributions are as follows:
\begin{itemize}
    \item We propose MASSV, a comprehensive framework that combines (1) a
    architectural adaptation connecting target VLM components with smaller
    language models from the same family, and (2) a self-data distillation
    technique specifically designed to align multimodal distributions for
    improved token acceptance.
    \item We provide extensive empirical evaluations demonstrating significant
    improvements in acceptance rates across multiple model families, with
    speedups reaching up to 1.28x overall on multimodal tasks.
    \item We present detailed ablations revealing that self-data distillation is
    crucial for multimodal drafting, improving distribution alignment between
    draft and target models particularly for visually-grounded tasks.
\end{itemize}

\section{Preliminaries}

We establish the necessary background for our approach. First, we review SD, an
inference acceleration technique that uses a smaller draft model to propose
tokens that are verified by a larger target model. Second, we  describe VLMs,
which combine visual encoders with language models to process multimodal inputs.
Finally, we discuss how SD has been adapted for VLMs, including the text-only
drafting baseline we compare against.

\begin{figure*}[!t]
\centering
\vskip -0.1in
\includegraphics[width=1.0\linewidth]{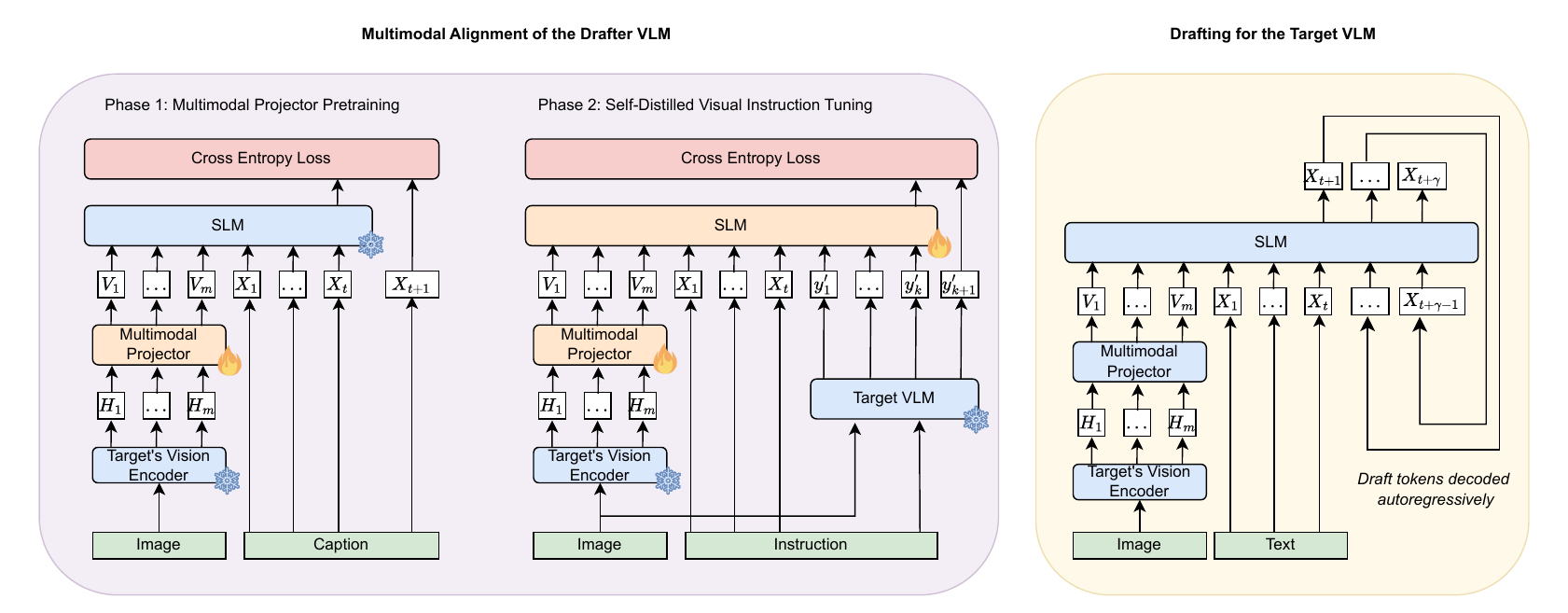}
\vskip -0.1in
\caption{Detailed architecture of MASSV illustrating: (1) the two-phase training methodology consisting of multimodal projector pretraining followed by self-distilled visual instruction tuning, and (2) the deployment configuration for draft token generation during speculative decoding inference. Components marked with the snowflake remain frozen during training to preserve their parameters, while components with the flame are trainable. This architecture enables efficient knowledge transfer from the target vision-language model to the smaller draft model while maintaining alignment in their token distributions.}
\label{fig:arch}
\end{figure*}

\subsection{Speculative Decoding}
\label{sec:speculative_decoding}
Speculative decoding~\citep{leviathan2023,
chen2023acceleratinglargelanguagemodel} is a technique for accelerating LLM
generation without altering the distribution of the generation output. In each
iteration of the algorithm, a draft model $M_q$ generates multiple draft tokens
that are verified in parallel by the target model $M_p$. The algorithm continues
iterating until an end-of-sequence (EOS) token is generated or the max sequence
length is reached.

Formally, let $X_{1:t} = X_1, X_2, ..., X_t$ be the input sequence for the
current iteration.~$M_q$ first autoregressively samples $\gamma$ draft tokens
$X_{t+1:t+\gamma}$, where token $X_{t+i}$ is sampled with probability $q(X_{t+i}
| X_{1:t+i-1})$. Next, $M_p$ computes the probabilities $p(X_{t+i} |
X_{1:t+i-1})$ for $i = 1,2,...,\gamma + 1$ in parallel with one forward call.
These probabilities are used to evaluate the draft tokens sequentially, with the
probability of accepting token $X_{t+i}$ being $\min\left(1, \frac{p(X_{t+i} |
X_{1:t+i-1})}{q(X_{t+i} | X_{1:t+i-1})}\right)$. If the token is accepted, it is
added to the generation output and the next token is evaluated. Otherwise, if
the token is rejected, a new token is sampled from the residual distribution
$\operatorname{norm}(\max(p(\cdot | X_{1:t+i-1}) - q(\cdot | X_{1:t+i-1}), 0))$
and the iteration ends. Sampling from the residual distribution ensures the
output distribution of the speculative decoding algorithm is the same as the
target's output distribution. 

In the degenerate case where sampling is disabled (temperature = 0), the
algorithm simplifies to greedy decoding. The draft model generates tokens by
selecting $X_{t+i} = \arg\max_{x} q(x | X_{1:t+i-1})$. During verification,
token $X_{t+i}$ is accepted if and only if $X_{t+i} = \arg\max_{x} p(x |
X_{1:t+i-1})$. If rejected, the token is set to $\arg\max_{x} p(x |
X_{1:t+i-1})$. 

\subsection{Vision-Language Models}
\label{sec:vlms}
Vision-language models (VLMs) process multimodal inputs, consisting of visual
and text tokens, by mapping the tokens into a joint embedding space. A VLM
consists of three components: a vision encoder $\phi_I$, multimodal projector
$g_\theta$, and a language model $M_p$. Given an input consisting of tokens
$X_{1:t}$ and visual information $I$, a VLM first extracts $m$ features $H_{1:m}
= \phi_I(I)$ from the image using the vision encoder. These image features are
then projected into the joint embeddings space $V_i = g_\theta(H_i)$ for $i \in
\{1,...,m\}$. Finally, the VLM samples the next token $X_{t+1}$ from
$p(\cdot|X_{1:t},V_{1:m})$, where $p(\cdot|\cdot)$ denotes the conditional
probability distribution of $M_p$. 

Note that directly using SD to accelerate a VLM on multimodal inputs requires
the drafter to also be a VLM. However,
~\citet{gagrani2024speculativedecodingmultimodallarge} show that a small
language model (SLM) can be used as an effective drafter by conditioning it only
on the text tokens in the input. Concretely, given an SLM drafter $M_q$, the
draft token $X_{t+i}$ is sampled from $q(\cdot|X_{1:t+i-1})$ for
$i=1,...,\gamma$. We refer to this as \textit{text-only drafting} and use it as
the baseline in our experiments.

\section{Methodology}
\label{sec:method}
We introduce a method to adapt an SLM into an effective draft model for
LLaVA-style vision-language models, which employ a modular architecture of
separate vision encoder and language model components connected via a projection
layer that maps image features into the language model's embedding space. Our
approach integrates the target VLM's frozen vision encoder into the SLM through
a randomly initialized MLP-based projector, preserving architectural
compatibility while enabling visual processing. We then align the adapted model
with the target VLM through a two-phase training protocol: (1) the projector is
pretrained on paired image-text data to establish robust visual grounding; and
(2) the model undergoes self-distilled visual instruction tuning to optimize
token-level distribution alignment. The overall architecture is illustrated in
Figure~\ref{fig:arch}.

\subsection{Architectural Adaptation}
Let $M_p^{\text{VLM}} = (\phi_I^p, g_\theta^p, M_p)$ denote the target VLM,
where $\phi_I^p$ is the vision encoder, $g_\theta^p$ is the multimodal
projector, and $M_p$ is the language model. Let $M_q$ be an SLM from the same
model family as $M_p$. While our method can be applied to any small language
model, this work specifically focuses on text-only SLMs from the same model
family as the larger VLM. This choice ensures that the draft model's tokenizer
and vocabulary are compatible with those of the target during SD. Although
recent work has demonstrated approaches to handle heterogeneous
vocabularies~\citep{timor2025acceleratingllminferencelossless}, these techniques
trade latency for vocabulary compatibility. Furthermore, existing methods have
not demonstrated their effectiveness in handling multiple modalities, as
required for VLMs. Due to these limitations and considerations beyond the scope
of this work, we leave exploring vocabulary heterogeneity in multimodal SD for
future research.

We construct the VLM drafter $M_q^{\text{VLM}}$ as follows,
\begin{equation}
M_q^{\text{VLM}} = (\phi_I^p, g_\psi^q, M_q),
\end{equation}
where $\phi_I^p$ is the shared vision encoder from the target VLM, $g_\psi^q$ is
a randomly initialized multimodal projector, and $M_q$ is the draft SLM. The
projector $g_\psi^q$ has the same architecture as $g_\theta^p$, but its output
dimension $d_{\text{out}}^q$ is set to match the embedding dimension of $M_q$,
\begin{equation}
g_\psi^q: \mathbb{R}^{d_{\text{vis}}} \rightarrow \mathbb{R}^{d_{\text{emb}}^q},
\end{equation}
where $d_{\text{vis}}$ is the vision encoder's output dimension and
$d_{\text{emb}}^q$ is the embedding dimension of $M_q$. We choose to share the
vision encoder between the target and the drafter, since this ensures that the
drafter and target process the same visual features $H_{1:m} = \phi_I^p(I)$ for
a given image input $I$. This architectural choice also reduces compute cost by
avoiding redundant vision encoding operations during inference.

\subsection{Multimodal Alignment}
\label{multimodal_alignment_section}
\textbf{Multimodal Projector Pretraining} Following~\citet{Liu_2024_CVPR}, we
first pretrain the multimodal projector $g_\psi^q$ by training the VLM drafter
with the vision encoder and SLM backbone frozen. Given a pretraining dataset
$D_{\text{pre}} = \{(I_j, C_j)\}_{j=1}^{N}$ of image-caption pairs, we optimize,
\begin{equation}
\mathcal{L}_{\text{pre}}(\psi) = - \sum_{j=1}^{N} \sum_{i=1}^{|C_j|} \log q_\psi(c_j^i | c_j^{1:i-1}, V_j),
\end{equation}
where $V_j = g_\psi^q(\phi_I^p(I_j))$ are the projected visual features, $c_j^i$
is the $i$-th token of caption $C_j$, and $q_\psi$ denotes the distribution of
the draft VLM with projector parameters $\psi$. Only $\psi$ is updated during
this phase while $\phi_I^p$ and $M_q$ remain frozen.

\textbf{Self-Distilled Visual Instruction Tuning}
In this phase, we introduce Self-Distilled Visual Instruction Tuning (SDViT), an
approach that employs self-data distillation (SDD) to align the drafter's
distribution with the target's multimodal distribution. Let $D = \{(I_i, X_i,
y_i)\}_{i=1}^n$ be a visual instruction dataset, where $I_i$ is the image input,
$X_i$ is the text instruction, and $y_i$ is the reference response.


The original SDD formulation
by~\citet{thangarasa2024selfdatadistillationrecoveringquality,
Yang2024SelfDistillationBD} generates target outputs using task-specific
contexts and templates. In contrast, for SD, our objective is to align the
drafter's token-level predictions with the target's. Therefore, we directly use
the target VLM to generate responses,
\begin{equation}
y_i^\prime = \mathrm{sample}_{\mathrm{top\text{-}p}}(p(\cdot | I_i, X_i)),
\end{equation}
where $p$ denotes the target VLM's distribution conditioned on both image $I_i$
and text instruction $X_i$. This creates a self-distilled dataset $D^\prime =
\{(I_i, X_i, y_i^\prime)\}_{i=1}^n$. We then fine-tune the drafter with its
vision encoder frozen to minimize,
\begin{equation}
\mathcal{L}_{\text{SDViT}}(\theta) = - \sum_{i=1}^{n} \sum_{k=1}^{|y_i^\prime|} \log q_\theta(y_i^{\prime k} | y_i^{\prime 1:k-1}, X_i, V_i),
\end{equation}
where $V_i = g_\psi^q(\phi_I^p(I_i))$ are the projected visual features,
$y_i^{\prime k}$ is the $k$-th token of the target's response, and $q_\theta$
denotes the drafter's distribution with parameters $\theta = \{\psi, \theta_q\}$
(projector and SLM parameters). 

In contrast to generic visual instruction tuning
with fixed dataset labels, our self-distillation strategy trains the drafter on
the target's actual outputs, directly optimizing for the acceptance mechanism in
SD. SDViT addresses this through diverse sampling, where the target VLM
generates responses across different temperature values with top-p sampling,
creating a varied dataset that better represents the full response distribution.
Specifically, draft tokens are accepted with probability $\min\left(1,
\frac{p(X_t|X_{1:t},I)}{q(X_t|X_{1:t},I)}\right)$. By training on the target's
outputs rather than generic labels, we maximize the overlap between the
drafter's distribution $q$ and the target's distribution $p$, leading to higher
token acceptance rates during inference. Our experimental results in
Section~\ref{ablation:sdd} show that this alignment translates to improved token
acceptance rates during SD.

The theoretical foundations of our approach are supported by recent findings on
distillation pitfalls. \citet{tiapkin2025teacherhackinglanguagemodel} identified
``teacher hacking'' as a key failure mode where students trained on fixed
offline datasets learn to exploit teacher imperfections  instead of
approximating the intended distribution. They found that data diversity is a
critical factor in preventing this phenomenon. Our self-data distillation
approach addresses this concern through the top-p sampling strategy described
earlier, which ensures varied responses from the target model. This diversity
prevents the drafter from overfitting to specific teacher patterns, explaining
why our approach achieves superior alignment compared to traditional
fixed-dataset distillation, particularly for visually-grounded tasks where
teacher hacking would be most problematic. Visual data presents unique
challenges for model distillation due to three key factors: higher
dimensionality of image features, difficulty in aligning visual and textual
representations, and inconsistent ways humans describe visual content in
language~\citep{huvpd}. These complexities make it particularly easy for
drafters trained on fixed datasets to learn superficial shortcuts rather than
developing genuine visual reasoning capabilities.

\begin{table*}[!h]
\caption{Mean accepted lengths ($\tau$) and speedups across model families,
tasks, and temperatures ($T \in \{0, 1\}$) with speculation length $\gamma = 5$.
Values show tokens accepted per target VLM forward pass, with speedup ratios in
parentheses (normalized to baseline). MASSV consistently outperforms the
text-only baseline~\citep{gagrani2024speculativedecodingmultimodallarge},
achieving substantial gains on visually-grounded tasks like COCO captioning
(+47.5\% at $T=0$: $2.21 \rightarrow 3.26$) and improving overall acceptance
(+30.1\% for Qwen2.5-VL 7B: $2.46 \rightarrow 3.20$). MASSV delivers practical
efficiency with $1.28\times$ end-to-end speedup for Qwen2.5-VL 7B at $T=0$ and
generalizes effectively to larger models without requiring direct alignment.}
\label{tab:final_results}
\begin{center}
\begin{small}
\begin{sc}
\begin{tabular}{lccccccccr}
\toprule
Target Model & Method & LLaVa 150k & LLaVA-Bench & GQA  & COCO & Overall \\
\midrule
\multicolumn{7}{c}{Temperature = 0}\\
\midrule
\multirow{2}{*}{\makecell[l]{Qwen2.5-VL 7B \\ Instruct}}  & Baseline & 2.37
(1.00x)                & 2.61 (1.00x)                             & 2.59 (1.00x)
& 2.21 (1.00x)             & 2.46 (1.00x)     \\
                       & MASSV     & 3.21 (1.24x)               & 3.12 (1.16x)
                       & 3.28 (1.26x) & 3.26 (1.46x)           &
                       $\textbf{3.20}_{\textcolor{OliveGreen}{\uparrow0.74}}$
                       ($\textbf{1.28x}$)    \\
\midrule
\multirow{2}{*}{\makecell[l]{Qwen2.5-VL 32B \\ Instruct}}  & Baseline & 2.46
(1.00x)                 & 2.70 (1.00x)                             & 2.79
(1.00x)  & 2.48 (1.00x)             & 2.61 (1.00x)     \\
& MASSV     & 3.12 (1.26x)               & 2.90 (1.07x)
& 3.19 (1.13x) & 3.09 (1.23x)           &
$\textbf{3.04}_{\textcolor{OliveGreen}{\uparrow0.43}}$ ($\textbf{1.17x}$)   \\
\midrule
\multirow{2}{*}{Gemma3-12B IT}         & Baseline & 2.71 (1.00x)
& 2.72 (1.00x)                             & 2.75 (1.00x)  & 2.84 (1.00x)
& 2.76 (1.00x)     \\
                       & MASSV     & 3.30 (1.19x)                 & 3.00 (1.11x)
                       & 3.07 (1.18x)  & 3.41  (1.24x)           &
                       $\textbf{3.19}_{\textcolor{OliveGreen}{\uparrow0.43}}$
                       (\textbf{1.18x})  \\ 
\midrule
\multirow{2}{*}{Gemma3-27B IT}  & Baseline & 2.49 (1.00x)                 & 2.70
(1.00x)                             & 2.61 (1.00x)  & 2.73 (1.00x)             &
2.65 (1.00x)     \\
& MASSV     & 3.00  (1.20x)                & 2.84  (1.05x)
& 2.86  (1.09x) & 3.24  (1.20x)             &
$\textbf{2.99}_{\textcolor{OliveGreen}{\uparrow0.34}}$  (\textbf{1.14x})    \\
\midrule
\multicolumn{7}{c}{Temperature = 1}\\
\midrule
\multirow{2}{*}{\makecell[l]{Qwen2.5-VL 7B \\ Instruct}} & Baseline & 2.47
(1.00x)                 & 2.75 (1.00x)                             & 2.63
(1.00x)  & 2.41 (1.00x)             & 2.58 (1.00x)     \\
                       & MASSV     & 3.35  (1.26x)               & 2.98 (1.09x)
                       & 3.19 (1.19x) & 3.31 (1.35x)           &
                       $\textbf{3.18}_{\textcolor{OliveGreen}{\uparrow0.60}}$
                       (\textbf{1.22x})\\
\midrule
\multirow{2}{*}{\makecell[l]{Qwen2.5-VL 32B \\ Instruct}}  & Baseline & 2.48
(1.00x)                 & 2.69 (1.00x)                            & 2.75 (1.00x)
& 2.56 (1.00x)            & 2.63 (1.00x)     \\
& MASSV     & 3.01 (1.25x)               & 2.87 (1.09x)
& 3.00 (1.09x) & 3.04 (1.19x)          &
$\textbf{2.97}_{\textcolor{OliveGreen}{\uparrow0.34}}$ (\textbf{1.15x})  \\
\midrule
\multirow{2}{*}{Gemma3-12B IT}        & Baseline & 2.67 (1.00x)                &
2.79  (1.00x)                           & 2.78 (1.00x) & 2.94 (1.00x)
& 2.82 (1.00x)    \\
                       & MASSV     & 3.08 (1.13x)               & 2.82 (1.05x)
                       & 3.01 (1.10x) & 3.37 (1.16x)           &
                       $\textbf{3.06}_{\textcolor{OliveGreen}{\uparrow0.24}}$
                       (\textbf{1.11x})   \\
\midrule
\multirow{2}{*}{Gemma3-27B IT}  & Baseline & 2.57 (1.00x)                & 2.67
(1.00x)                            & 2.63 (1.00x) & 2.73 (1.00x)            &
2.67 (1.00x)     \\
& MASSV     & 2.81  (1.09x)              & 2.62 (1.02x)
& 2.82 (1.07x) & 3.13 (1.15x)           &
$\textbf{2.84}_{\textcolor{OliveGreen}{\uparrow0.17}}$ (\textbf{1.08x})  \\
\bottomrule
\end{tabular}
\end{sc}
\end{small}
\end{center}
\end{table*}

\section{Empirical Results} 
\label{sec:results}
We organize this section into two subsections, clearly separating the
experimental setup and detailed results.

\subsection{Experimental Setup}
\label{sec:experimental_setup}
\textbf{Draft and Target Models.} Our evaluation leverages two distinct model
families: the Qwen2.5-VL Instruct~\citep{bai2025qwen25vltechnicalreport} and
instruction-tuned Gemma3~\citep{gemmateam2025gemma3technicalreport}.
Specifically, for Qwen2.5-VL, we set the 7B model as our primary target,
applying MASSV to Qwen2.5-1.5B Instruct. Similarly, for Gemma3, we target the
12B IT variant and adapt Gemma3-1B IT using MASSV. We selected these specific
SLMs because they are from the same model families as the larger target models
and were readily available as checkpoints on HuggingFace. We utilize
\textit{text-only drafting} with the off-the-shelf SLM as our baseline (1.00x).


\textbf{Drafter Training for Multimodal Adaptation.} The draft model training
process consists of two distinct phases and requires only moderate compute
infrastructure, achievable with standard research hardware (e.g., four-GPU
server with current-generation accelerators). Initially, we pretrain each
drafter for one epoch on the
LLaVA-Pretrain-LCS-558K~\footnote{\href{https://huggingface.co/datasets/liuhaotian/LLaVA-Pretrain}{https://huggingface.co/datasets/liuhaotian/LLaVA-Pretrain}}
dataset, using a global batch size of 256 and a learning rate of 1 x
10$^{\text{-4}}$. Subsequently, we fine-tune the models on data distilled from
the
LLaVA-mix-665K~\footnote{\href{https://huggingface.co/datasets/liuhaotian/LLaVA-Instruct-150K/blob/main/llava_v1_5_mix665k.json}{https://huggingface.co/datasets/liuhaotian/LLaVA-Instruct-150K/blob/main/llava\_v1\_5\_mix665k.json}}
dataset for another epoch with a batch size of 128 and learning rate 2 x
10$^{\text{-5}}$. See Appendix~\ref{app:experimental} for more details.

\textbf{Evaluation Tasks.} We conduct evaluations using four multimodal
benchmarks: LLaVA Instruct 150k~\citep{liu2023visual}, LLaVA-Bench
(In-the-Wild)~\footnote{\href{https://huggingface.co/datasets/liuhaotian/llava-bench-in-the-wild}{https://huggingface.co/datasets/liuhaotian/llava-bench-in-the-wild}},
GQA~\citep{Hudson_2019_CVPR}, and image captioning prompts from COCO Test
2017~\citep{lin2015microsoftcococommonobjects}. Performance is measured by mean
accepted length ($\tau$), which quantifies the average number of tokens accepted
per forward pass of the target model, directly correlating to speedup
independent of hardware. Evaluation prompts for GQA reasoning and COCO
Captioning tasks are provided in Appendix~\ref{app:eval}.

\textbf{Inference Settings.} During inference, all drafters run on a single H100
GPU, with speculation length set to $\gamma = 5$. We evaluate performance at
sampling temperatures $T \in \{0, 1\}$.

\begin{figure*}[!t]
    \centering
    \vskip -0.1in
    \includegraphics[width=0.9\linewidth]{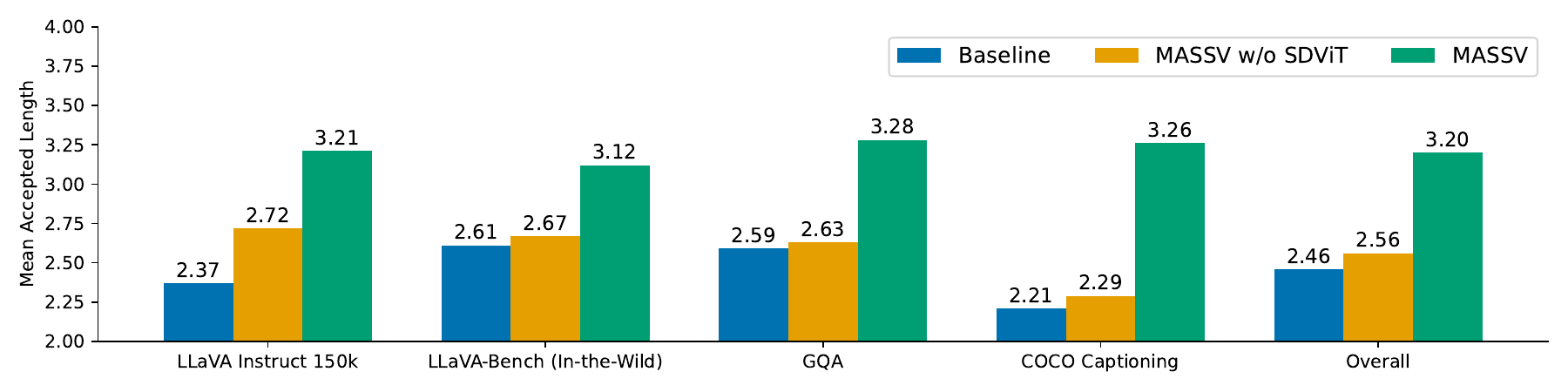}
    \vskip -0.1in
    \caption{Mean accepted lengths when drafting for Qwen2.5-VL 7B Instruct at temperature $T = 0$ with speculation length $\gamma = 5$. The baseline uses Qwen2.5-1.5B as a text-only drafter (image tokens removed). MASSV achieves a substantial improvement in token acceptance across all tasks, increasing overall mean accepted length from 2.46 to 3.20 (+30.1\%).}
    \label{fig:qwen_mal}
\end{figure*}

\subsection{Results}
Our results demonstrate MASSV's significant improvements over the text-only
baseline across all evaluated settings (see Table~\ref{tab:final_results}). At
temperature $T = 0$, MASSV achieves a noticeable increase in mean accepted
length (MAL), most notably improving by $30.1\%$ (from 2.46 to 3.20) for the
Qwen2.5-VL 7B Instruct model. Similarly, at $T = 1$, MASSV attains a MAL
improvement of $23.3\%$ (from 2.58 to 3.18). These improvements are consistent
across different downstream tasks, with the largest relative gains observed in
visually intensive tasks such as COCO captioning. For instance, MASSV increases
MAL by $47.5\%$ (2.21 to 3.26) on COCO captioning tasks at $T = 0$, highlighting
the importance of multimodal drafting for visually-grounded generations.
Moreover, MASSV consistently outperforms the baseline on the Gemma3 family
despite their significant architectural differences (e.g., dynamic visual token
count in Qwen2.5-VL versus interleaved sliding window attention in Gemma3).
Specifically, MASSV improves MAL by $15.6\%$ (2.76 to 3.19) on Gemma3-12B IT at
$T = 0$, demonstrating its effectiveness across diverse VLMs.
\vspace{-5pt}
\paragraph{Generalization to Larger Model Variants.} We also evaluated MASSV on
larger variants within each model family, specifically Qwen2.5-VL 32B and
Gemma3-27B. Although we did not directly apply SDViT to these larger targets, we
hypothesized that MASSV, when applied to smaller distilled versions (7B and
12B), could still benefit their larger counterparts due to their shared
architecture and distillation lineage. Our empirical results confirm this
hypothesis, demonstrating that MASSV provides meaningful gains even when scaling
up within the same model family. This finding is particularly impactful as it
allows substantial computational and time savings by enabling MASSV adaptation
on smaller, more efficient targets, which can subsequently generalize to larger
models. We leave a thorough exploration of this promising direction for future
research.

In summary, our experiments show MASSV's effectiveness in closely aligning draft
predictions with multimodal target models. MASSV consistently outperforms
text-only drafting baselines, demonstrating strong generalization across diverse
architectures, tasks, and importantly, larger-scale model variants. This
generalization provides significant practical advantages, notably computational
efficiency, by allowing MASSV alignment with smaller target models, which
subsequently transfers performance improvements to larger models from the same
family without additional training. These findings highlight MASSV as a
promising and scalable solution for multimodal speculative decoding.
\vspace{-5pt}
\paragraph{End-to-end Inference Speedups.} The mean accepted length improvements
translate directly to substantial wall-clock speedups during inference. MASSV
achieves an overall end-to-end speedup of 1.28$\times$ for Qwen2.5-VL 7B
Instruct at temperature $T = 0$, with even higher speedups on specific tasks
such as COCO captioning (1.46$\times$). These speedups remain consistent across
model families, with Gemma3-12B IT achieving 1.18$\times$ acceleration. Notably,
MASSV demonstrates effective scalability to larger models, achieving
1.17$\times$ speedup for Qwen2.5-VL 32B and 1.14$\times$ for Gemma3-27B, despite
not requiring direct alignment on these larger targets. These results show that
MASSV's improved token acceptance rates translate to meaningful practical
efficiency gains across diverse model architectures.



\section{Ablation Studies}
We investigate the critical components of our approach through two ablation
studies. First, we evaluate the impact of self-distilled visual instruction
tuning on distribution alignment. Second, we examine whether multimodal
capability provides meaningful benefits over text-only drafting.

\subsection{Effect of Self-Distilled Visual Instruction Tuning}
\label{ablation:sdd}
We assess the role of self-distilled distillation in our method by comparing
drafters trained with SDViT versus standard fine-tuning on a vanilla dataset.
Specifically, we adapt Qwen2.5-1.5B Instruct and Gemma3-1B IT into drafters for
Qwen2.5-VL 7B Instruct and Gemma3-12B IT, respectively.
Figure~\ref{fig:qwen_mal} demonstrates the efficacy of MASSV with SDViT (green
bar) for Qwen2.5-VL 7B Instruct across diverse multimodal benchmarks. MASSV
exhibits substantial performance gains, most prominently in COCO Captioning
where the mean accepted length increases from 2.21 to 3.26 tokens (+47.5\%).
Table~\ref{tab:sdd-ablation-table} summarizes our comprehensive ablation study
on SDViT across both target models: Qwen2.5-VL 7B Instruct and Gemma3-12B IT.
The quantitative evaluation results clearly demonstrate the critical importance
of self-distilled visual instruction tuning for effective multimodal SD. For the Gemma3 architecture, without SDViT (denoted as
MASSV$_\text{w/o SDViT}$), the Gemma3-1B IT draft model exhibits a significant
performance regression, with mean accepted length deteriorating to 2.33 compared
to the baseline's 2.74 (a 13\% decrease in acceptance rate). This indicates that
naive architectural adaptation without distribution alignment can be notably
detrimental to performance. In contrast, when enhanced with SDViT, the model
achieves a mean accepted length of 3.14, representing a substantial 14.6\%
improvement over the baseline and a 1.18x speedup. These results highlight the
critical role of distribution alignment in multimodal SD.

\begin{table}[!t]
\caption{Ablation results on the effect of SDViT on drafting performance. Qwen2.5-1.5B Instruct and Gemma3-1B IT are the base SLMs used to create drafters for Qwen2.5-VL 7B Instruct and Gemma3-12B IT, respectively. The reported mean accepted lengths ($\tau$) are measured on the overall multimodal speculative decoding benchmark dataset at temperature = 0.}
\label{tab:sdd-ablation-table}
\begin{center}
\begin{small}
\begin{sc}
\begin{tabular}{lccc}
\toprule
Target & Method & $\tau$ & Speedup \\
\midrule
& Baseline & 2.46 & 1.00x \\
\makecell[l]{Qwen2.5-VL 7B \\ Instruct} & MASSV$_\text{w/o SDViT}$ & 2.56 &
1.04x \\
& MASSV & 3.20 & 1.29x \\
\midrule
& Baseline & 2.74 & 1.00x \\
Gemma3-12B IT & MASSV$_\text{w/o SDViT}$ & 2.33 & 0.87x \\
& MASSV & 3.14 & 1.18x \\
\bottomrule
\end{tabular}
\end{sc}
\end{small}
\end{center}
\vskip -0.1in
\end{table}

\begin{figure}[!t]
\centering
\vskip -0.1in
\includegraphics[width=1\linewidth]{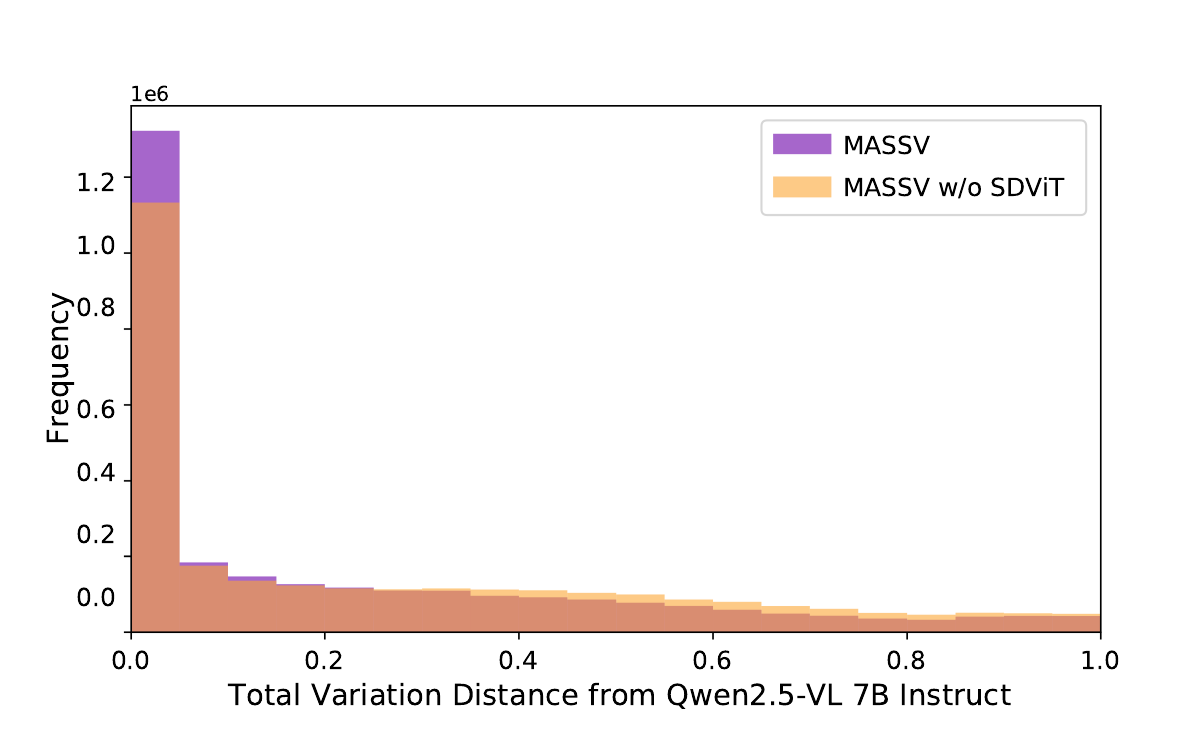}
\vskip -0.1in \caption{Histogram of total variation distances (TVD), comparing
the Qwen2.5-1.5B drafters trained with (purple) and without (orange)
self-distilled visional instruction (SDViT) against the Qwen2.5-VL 7B target
model on our multimodal SD benchmark. MASSV yields a highly
skewed distribution concentrated at low TVD values, indicating tighter alignment
with the target’s token distribution. In contrast, MASSV$_{\text{w/o SDViT}}$
produces a broader, heavier-tailed distribution, reflecting reduced alignment.
The left-skewed shape of the MASSV distribution quantitatively suggests that
SDViT narrows the distributional gap between draft and target.}
\label{fig:tvd}
\end{figure}

\vspace{-5pt}
\paragraph{Distribution Analysis.} To understand the mechanism behind these
improvements, we analyze the distribution alignment between drafters and
targets. For each multimodal input, we compute the Total Variation Distance
(TVD) between the drafter's and target's output token distributions. The TVD
measures the maximum difference between two probability distributions,
\begin{equation}
\label{eq:tvd}
\text{TVD}(P, Q) = \frac{1}{2} \sum_{x \in \mathcal{X}} |P(x) - Q(x)|,
\end{equation}
where $P$ and $Q$ are the target and drafter token distributions, respectively,
and $\mathcal{X}$ is the vocabulary. TVD is particularly relevant in the context
of SD, as it bounds the expected probability that tokens
proposed by the draft model will be rejected by the target model. By minimizing
TVD through our SDViT approach, we directly optimize for higher token acceptance
rates, which explains the improved mean accepted length observed in our
experiments. For discrete distributions like token probabilities, TVD ranges
from 0 (identical distributions) to 1 (completely disjoint distributions).
Figure~\ref{fig:tvd} shows the resulting distribution. 
The drafter trained with SDViT produces significantly more tokens with output
distributions closely matching the target. This demonstrates that SDViT enables
the drafter to more faithfully reproduce the target model's token-level
behavior. These results indicate that: (1) self-data distillation substantially
improves distribution alignment between drafter and target, and (2) distribution
alignment contributes more to drafting performance than raw multimodal
capability.

\subsection{Text-Only vs Multimodal Drafting}
Given that distribution alignment appears more important than multimodal
capability, we investigate whether multimodal processing provides any benefit
over text-only drafting. This question is particularly relevant since text-only
drafting could offer computational advantages by avoiding visual encoding
operations during the draft phase.

We evaluate our VLM drafters in text-only mode by discarding visual tokens from
the input, thereby using only the language model backbone of our adapted
drafter. This approach mirrors the baseline strategy used in prior
work~\citep{gagrani2024speculativedecodingmultimodallarge}, where standard SLMs
trained from scratch serve as drafters for VLM targets without processing any
visual information. Table~\ref{tab:textonly-ablation-table} shows that
multimodal drafting consistently outperforms text-only drafting across both
model families. The improvements are substantial: 12.7\% higher mean accepted
length for Qwen2.5-VL (3.20 vs.~2.84) and 6.7\% higher for Gemma3 (3.19
vs.~2.99). These gains demonstrate that while distribution alignment is the
primary factor in drafting performance, incorporating visual information
provides additional benefits for predicting the target VLM's outputs.

\begin{table}[!t]
\caption{Ablation results on the performance of text-only drafting. The VLM drafter's language model backbone serves as a text-only drafter by discarding all visual tokens. Mean accepted lengths ($\tau$) are measured on the overall benchmark dataset at temperature = 0.}
\label{tab:textonly-ablation-table}
\begin{center}
\begin{small}
\begin{sc}
\begin{tabular}{lcc}
\toprule
Target Model & Method & $\tau$ \\
\midrule
\multirow{2}{*}{\makecell[l]{Qwen2.5-VL 7B \\ Instruct}} & Text-Only & 2.84 \\
& Multimodal & 3.20 \\
\midrule
\multirow{2}{*}{Gemma3-12B IT} & Text-Only & 2.99 \\
& Multimodal & 3.19 \\
\bottomrule
\end{tabular}
\end{sc}
\end{small}
\end{center}
\vskip -0.1in
\end{table}

The advantage of multimodal drafting likely stems from its ability to condition
token predictions on the actual visual content, particularly for
visually-grounded tokens such as object names, spatial relationships, and visual
attributes. While text-only drafting must rely solely on linguistic patterns and
context, multimodal drafting can leverage direct visual evidence to better
predict the target VLM's outputs.

Based on these observations, we focus exclusively on multimodal drafting in our
main experiments (Section~\ref{sec:results}). This choice ensures we capture the
full benefits of visual information while maintaining strong distribution
alignment through SDViT. As we demonstrate
across multiple model families and tasks, this combination of multimodal
capability and distribution alignment yields consistent improvements in
SD performance.

\section{Related Work}

Speculative decoding has emerged as a promising technique for accelerating LLM
inference without compromising output quality. This approach leverages smaller,
faster draft models to autoregressively generate multiple candidate tokens,
which are then verified in parallel by the larger target model in a single
forward pass~\citep{leviathan2023, chen2023acceleratinglargelanguagemodel}. The
theoretical foundations of this technique were established by identifying
conditions under which speculative proposals can preserve the original model's
output distribution~\citep{leviathan2023}. Recent advancements include
tree-structured variants~\citep{li2024eaglespeculativesamplingrequires,
li2024eagle2fasterinferencelanguage, wang2025opttreespeculativedecodingadaptive,
chen2024sequoia}, self-drafting~\citep{elhoushi-etal-2024-layerskip,
zhang-etal-2024-draft, liu2024kangaroo, xia2025swift},
N-gram-based~\citep{stewart2024ngrammysacceleratingautoregressiveinference,
ou2024losslessaccelerationlargelanguage} and
retrieval-based~\citep{he-etal-2024-rest,
yang2023inferencereferencelosslessacceleration} that further enhance inference
efficiency. However, these approaches have primarily focused on text-only
models, where the draft and target  operate within the same modality space.

\vspace{-5pt}
\paragraph{Multimodal Speculative Decoding.} 
Extending speculative decoding to vision-language models introduces fundamental
challenges absent in unimodal settings.
\citet{gagrani2024speculativedecodingmultimodallarge} conducted initial
explorations in this domain by evaluating several draft model variants with the
LLaVA-7B architecture~\citep{Liu_2024_CVPR}. They systematically investigated
four 115M-parameter draft models: (1) a base model pre-trained on 600B tokens,
(2) an instruction-tuned chat model, (3) a text-only variant fine-tuned on LLaVA
data, and (4) a multimodal version incorporating a subcloned image adapter from
the target model. Their analysis across image question-answering, captioning,
and reasoning tasks revealed modest token acceptance rates, with the multimodal
variant achieving only marginal improvements over text-only counterparts despite
added computational complexity. Detailed traces demonstrated that while drafters
successfully predicted function words and repeated tokens, they struggled with
visually-grounded content, highlighting two fundamental challenges: (1)
architectural misalignment between drafters and vision-language targets, and (2)
distributional divergence between text-only priors and visually-informed
outputs. \citet{lee2024inbatch} introduced a batch-based approach that combines
predictions from multiple drafting methods to increase the likelihood of tokens
passing verification. While their ensemble technique improves empirical
performance without parameter overhead, it operates primarily as a post-hoc
aggregation mechanism rather than addressing the underlying distributional
divergence between individual drafters and the target model. Our self-distilled
multimodal drafting framework directly addresses these limitations through
principled vision-language alignment techniques.

\vspace{-5pt}
\paragraph{Draft Model Alignment.} Self-distillation uses a model’s own outputs
as training targets, extending traditional knowledge distillation approaches.
While~\citet{Yang2024SelfDistillationBD} showed self-distillation can bridge
distribution gaps during language model fine-tuning
and~\citet{thangarasa2024selfdatadistillationrecoveringquality} demonstrated its
effectiveness in mitigating catastrophic forgetting in pruned models, we extend
these insights to multimodal drafting. In particular,
SD$^2$~\citep{lasby2025sd2selfdistilledsparsedrafters} apply self-data
distillation to fine-grained sparse draft models, aligning them closely with
their original dense counterparts and yielding substantially higher mean
accepted lengths than undistilled sparse drafters. Unlike previous work, we
explicitly formulate self-distillation as an optimization for token acceptance
probability in the speculative decoding framework. By training our draft model
on responses generated by the target VLM itself rather than fixed dataset
labels, we align the draft model’s distribution with that of the target model.
This approach creates a direct optimization path that maximizes the likelihood
of draft tokens being accepted during inference. The result is drafters that
effectively capture the target model’s visual reasoning capabilities without
requiring extensive pre-training or architectural modifications. Our method
particularly improves acceptance rates for tokens related to visual content,
addressing a key limitation in prior multimodal drafting approaches.

\section{Conclusion}
In this paper, we present MASSV, a method to transform smaller language-only
models into highly efficient speculative drafters for vision-language models
(VLMs). MASSV addresses challenges like architectural incompatibility and
distribution mismatch by grafting the frozen vision encoder of the target VLM
onto the draft model via a trainable projector and aligning the drafter's token
distribution through fine-tuning on self-generated vision-language data. Across
both Qwen2.5-VL and Gemma3 model families, MASSV increases mean accepted length
by 16–30\% with end-to-end inference speedups of up to 1.46x, without
compromising output quality. Ablation studies underscore that self-data
distillation is crucial for distribution alignment, and full multimodal drafting
consistently outperforms text-only approaches, particularly for visually
grounded tasks. Given its generalizability and demonstrated performance gains,
MASSV presents a readily deployable solution for significantly accelerating VLM
inference across diverse architectures and tasks, unlocking the potential for
more efficient and responsive multimodal applications.


\section{Acknowledgements}
We would like to express our sincere gratitude to Eric Sather and Mike Lasby for
their thoughtful feedback and fruitful discussions that helped improve the final
manuscript. We also acknowledge Rohan Gupta and Eugene Osovetsky for their
valuable insights which emphasized the critical importance of ML
software/hardware co-design in addressing these complex challenges.

\bibliography{refs}
\bibliographystyle{icml2025}

\newpage
\appendix
\onecolumn
\section{Additional Experimental Details}
\label{app:experimental}
\begin{figure}[h]
\begin{subfigure}{0.5\textwidth}
\includegraphics[width=0.9\linewidth, height=6cm]{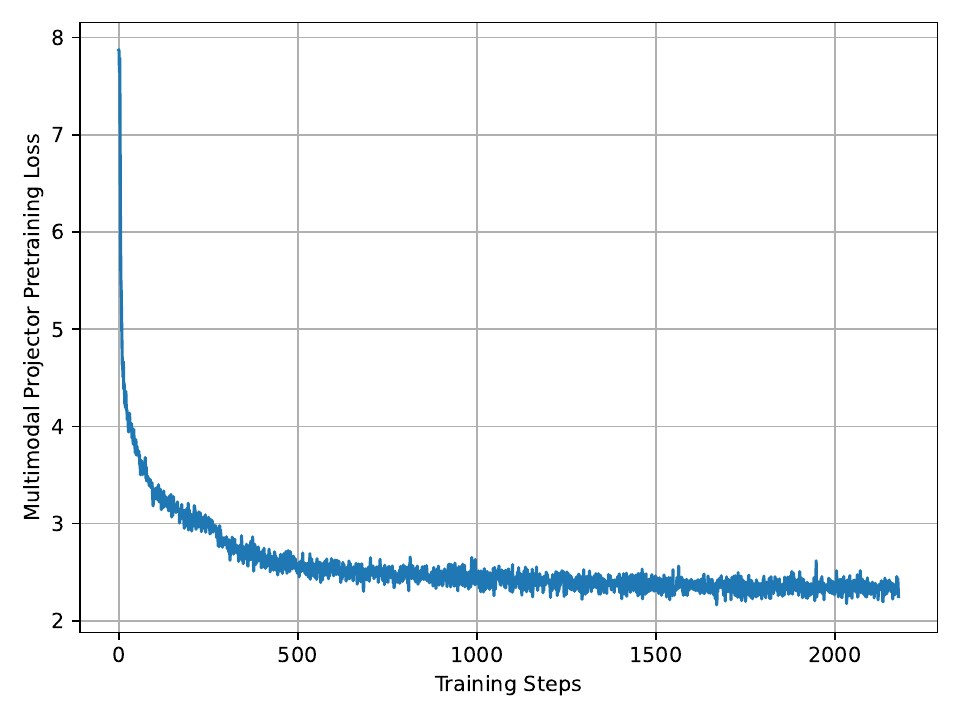} 
\caption{Phase 1: Multimodal Alignment}
\label{fig:subim1}
\end{subfigure}
\begin{subfigure}{0.5\textwidth}
\includegraphics[width=0.9\linewidth, height=6cm]{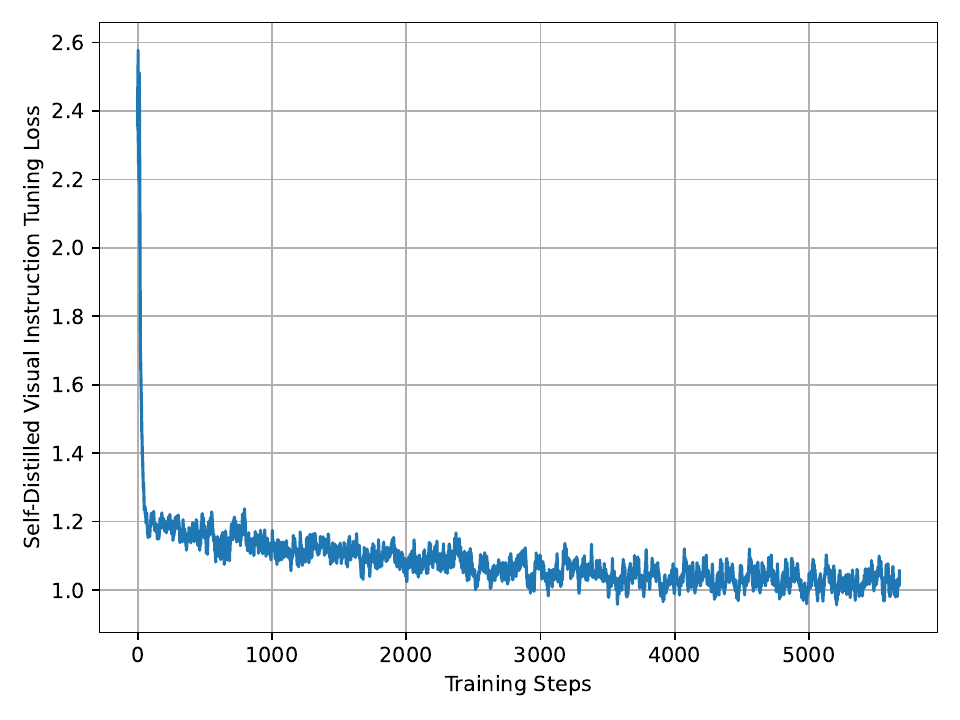}
\caption{Phase 2: Self-Distilled Visual Instruction Tuning}
\label{fig:subim2}
\end{subfigure}
\caption{Training loss curves obtained during the two-phase MASSV training process when adapting Qwen2.5-1.5B Instruct into a VLM drafter for Qwen2.5-VL 7B Instruct. (a) shows the cross-entropy loss during multimodal projector pretraining, which rapidly decreases from $\sim$8.0 to $\sim$2.5 within 2000 steps, indicating efficient adaptation of the trainable projector. (b) displays the loss trajectory during fine-tuning with self-generated target VLM responses, with stable convergence around 1.1 across 5000 training steps, demonstrating successful token distribution alignment between the draft and target models.}
\label{fig:training_curves}
\end{figure}

The training curves presented in Figure~\ref{fig:training_curves} of illustrate
the convergence patterns for both phases of the MASSV methodology described in
Section~\ref{sec:method}. In Phase~1 (Multimodal Alignment), the multimodal
projector pretraining loss exhibits rapid convergence within the first 500
steps, starting from approximately 8.0 and stabilizing around 2.5 by step 2000.
This demonstrates effective knowledge transfer from the target VLM's vision
encoder to the draft model via the trainable projector. Phase~2 (Self-Distilled
Visual Instruction Tuning) shows a more gradual optimization process with the
loss starting at approximately 2.6 and stabilizing around 1.1 with minor
fluctuations across 5000 training steps. These training dynamics align with our
experimental setup where each drafter was first pretrained for one epoch on the
LLaVA-Pretrain-LCS-558K dataset (batch size 256, learning rate $10^{-3}$),
followed by fine-tuning on data distilled from LLaVA-mix-665K (batch size 128,
learning rate $2 \times 10^{-5}$) using the target VLM. The convergence patterns
show successful training of both the multimodal projector and subsequent
distribution alignment through self-distilled visual instruction tuning.

\section{Evaluation}
\label{app:eval}

The following prompt templates were used during the evaluations described in
Section~\ref{sec:experimental_setup}. The GQA prompt explicitly requests
reasoning explanations alongside answers, evaluating the model's visual
reasoning capabilities. The COCO Captioning prompt elicits detailed image
descriptions without stylistic constraints. These standardized prompts ensure
consistent evaluation across all model variants (baseline, MASSV without SDViT,
and full MASSV), enabling fair comparison of mean accepted length and end-to-end
speedup metrics. By maintaining these consistent prompt templates, we facilitate
meaningful performance comparison not only within our experimental framework but
also with previously published results in multimodal speculative decoding
research. 

\begin{tcolorbox}[colback=lightgray!30, colframe=lightgray!30, width=\textwidth, arc=0mm, boxrule=0pt]
\textbf{Prompt for GQA Evaluation}

For the following question, provide a detailed explanation of your reasoning
process. Please analyze the visual elements systematically and articulate each
step of your thought process leading to the final answer. $\{\{Question\}\}$
\end{tcolorbox}

\begin{tcolorbox}[colback=lightgray!30, colframe=lightgray!30, width=\textwidth, arc=0mm, boxrule=0pt]
\textbf{Prompt for COCO Captioning Evaluation}

Examine the provided image carefully and generate a comprehensive description.
Please include relevant details about objects, their spatial relationships,
activities, attributes, and any other notable visual elements.
\end{tcolorbox}

\end{document}